\documentclass{article}

\usepackage[preprint]{neurips_2023}

\usepackage[utf8]{inputenc} 
\usepackage[T1]{fontenc}    
\usepackage{hyperref}       
\usepackage{url}            
\usepackage{booktabs}       
\usepackage{amsfonts}       
\usepackage{nicefrac}       
\usepackage{microtype}      
\usepackage{xcolor}         
\usepackage{soul}
\usepackage{tablefootnote}
\usepackage[normalem]{ulem}
\usepackage{multirow}
\usepackage{makecell}
\usepackage{enumitem}
\usepackage{adjustbox}
\usepackage{amsmath}
\usepackage{array}
\usepackage{graphicx}
\usepackage{comment}
\usepackage{booktabs}
\usepackage{xcolor,colortbl} 
\usepackage{algorithm}

\usepackage{bbm}
\usepackage{tabularx}
\usepackage{hyperref}

\usepackage{graphicx}
\usepackage{subcaption}
\usepackage{arydshln}
\usepackage{array}
\usepackage{booktabs}
\usepackage{pifont}
\usepackage{cleveref}
\crefname{section}{§}{§§}
\usepackage{amsthm}
\usepackage{todonotes}
\usepackage{color}
\usepackage{amsmath, bm, amsfonts}
\usepackage{newfloat}
\usepackage{listings}
\usepackage{multirow}
\usepackage{xspace}
\newcommand{\method}{{MemoryBank}\xspace}
\newcommand{\chatbot}{{SiliconFriend}\xspace}

\newcommand{\chatbotbelle}{\chatbot\textsubscript{BELLE}\xspace}
\newcommand{\chatbotchatglm}{\chatbot\textsubscript{ChatGLM}\xspace}
\newcommand{\chatbotchatgpt}{\chatbot\textsubscript{ChatGPT}\xspace}

\newcommand{\sect}{\S\xspace}
\newcommand{\saveSpaceSec}{\vspace{0pt}}

\title{MemoryBank: Enhancing Large Language Models with Long-Term Memory}

\author{
	Wanjun Zhong$^{1}$, Lianghong Guo$^{1}$, Qiqi Gao$^{2}$, He Ye$^{3}$,  Yanlin Wang$^{1}$\\
	$^1$ Sun Yat-Sen University 
	$^2$ Harbin Institute of Technology \\
	$^3$ KTH Royal Institute of Technology \\
	\texttt{\{zhongwj25@mail2, wangylin36@mail\}.sysu.edu.com} \\ 
	\texttt{2231612405@qq.com, 18b903026@stu.hit.edu.cn}\\
	\texttt{heye@kth.se}
}

\begin{document}

\maketitle

\begin{abstract}

Revolutionary advancements in Large Language Models (LLMs) have drastically reshaped our interactions with artificial intelligence (AI) systems, showcasing impressive performance across an extensive array of tasks. Despite this, a notable hindrance remains—the deficiency of a long-term memory mechanism within these models. This shortfall becomes increasingly evident in situations demanding sustained interaction, such as personal companion systems, psychological counseling, and secretarial assistance.
Recognizing the necessity for long-term memory, we propose MemoryBank, a novel memory mechanism tailored for LLMs. MemoryBank enables the models to summon relevant memories, continually evolve through continuous memory updates, comprehend, and adapt to a user's personality over time by synthesizing information from previous interactions. 
To mimic anthropomorphic behaviors and selectively preserve memory, MemoryBank incorporates a memory updating mechanism, inspired by the Ebbinghaus Forgetting Curve theory. This mechanism permits the AI to forget and reinforce memory based on time elapsed and the relative significance of the memory, thereby offering a more human-like memory mechanism and enriched user experience.
MemoryBank is versatile in accommodating both closed-source models like ChatGPT and open-source models such as ChatGLM.
To validate MemoryBank's effectiveness, we exemplify its application through the creation of an LLM-based chatbot named SiliconFriend in a long-term AI Companion scenario. Further tuned with psychological dialog data, SiliconFriend displays heightened empathy and discernment in its interactions.
 Experiment involves both qualitative analysis with real-world user dialogs and quantitative analysis with simulated dialogs. 
 In the latter, ChatGPT acts as multiple users with diverse characteristics and generates long-term dialog contexts covering a wide array of topics. The results of our analysis reveal that SiliconFriend, equipped with MemoryBank, exhibits a strong capability for long-term companionship as it can provide emphatic response, recall relevant memories and understand user personality. This underscores the effectiveness of MemoryBank\footnote{The materials are released in \url{https://github.com/zhongwanjun/MemoryBank-SiliconFriend}. The corresponding author is Yanlin Wang.}.
 
\end{abstract}

\saveSpaceSec
\section{Introduction}
\saveSpaceSec
The advent of Large Language Models (LLMs) such as ChatGPT
~\citep{chatgpt} and GPT-4~\citep{openai2023gpt4} has led to increasing influence across various sectors, from education and healthcare to customer service and entertainment. 
These powerful AI systems have demonstrated a remarkable ability to understand and generate human-like responses. Despite the remarkable capabilities of LLMs, a key limitation is their lack of long-term memory, an essential aspect of human-like communication, particularly noticeable in scenarios requiring sustained interactions like personal companionship, psychological counseling, and secretarial tasks.
Long-term memory in AI is vital to maintain contextual understanding, ensure meaningful interactions and understand user behaviors over time. For instance, personal AI companions need to recall past conversations for rapport building. In psychological counseling, an AI can provide more effective support with knowledge of the user's history and past emotional states. Similarly, secretarial AI requires memory for task management and preference recognition. The absence of long-term memory in LLMs hinders their performance and user experience. Therefore, it is essential to develop AI systems with improved memory capabilities for a more seamless and personalized interaction.

Therefore, we introduce \method, a novel mechanism designed to provide LLMs with the ability to retain long-term memory and draw user portraits. \method enables LLMs to recall historical interactions, continually evolve their understanding of context, and adapt to a user's personality based on past interactions, thereby enhancing their performance in long-term interaction scenarios.
Inspired by the Ebbinghaus Forgetting Curve theory, a well-established psychological principle that describes how the strength of memory decreases over time, \method further incorporates a dynamic memory mechanism closely mirroring human cognitive process. This mechanism empowers the AI to remember, selectively forget, and strengthen memories based on time elapsed, offering more natural and engaging user experience. 
Specifically, \method is built on a memory storage with memory retrieval and updating mechanism, and ability to summarize past events and users' personality.

\method is versatile as it can accommodate both closed-source LLMs like ChatGPT and open-source LLMs like ChatGLM~\citep{zeng2022glm} or BELLE~\citep{BELLE}.

To exemplify the practical implications of \method, we develop \chatbot, an LLM-based AI Companion chatbot integrated with this innovative memory mechanism. 
\chatbot is designed to retain and reference past interactions, reinforcing the transformative influence of \method in crafting a more personable AI companion.
A distinctive features of SiliconFriend is its tuning with 38k psychological conversations, collected from various online sources, which enables it to exhibit empathy, carefulness, and provide useful guidance, making it adept at handling emotionally charged dialogues. 
Moreover, one of the standout capabilities of SiliconFriend is to understand a user's personality by summarizing from past interactions, which empowers it to tailor responses to the user's individual traits, thereby enhancing user experience.
Additionally, SiliconFriend supports bilingual functionality, catering to users who communicate in English and Chinese. This multi-language support broadens its accessibility and usability across different user groups. 
SiliconFriend is implemented with two open-source models, ChatGLM and BELLE, along with one closed-source model, ChatGPT, showcasing the versatility of \method in accommodating different LLMs.

To evaluate the effectiveness of \method, we conduct evaluations covering both qualitative and quantitative analyses, where the former involves real-world user dialogs and the latter employs simulated dialogs.
For the quantitative analysis, we create a memory storage consisting of 10 days of conversations encompassing a diverse range of topics. These conversations involve 15 virtual users with diverse personalities, for which ChatGPT plays the role of users and generates dialog contexts according to their personalities. 
Based on this memory storage, we design 194 probing questions to assess whether the model could successfully recall pertinent memories and provide appropriate responses.
Experiment results showcase the capabilities of \chatbot in memory recall, provision of empathetic companionship, and understanding of user portraits. These findings corroborate the potential of \method to significantly improve the performance of LLMs in long-term interaction scenarios.
In this paper, we summarize the key contributions as follows:
\begin{itemize} [topsep=0pt,itemsep=0pt,partopsep=0pt,parsep=0pt,leftmargin=20pt]
    \item We introduce \method, a novel human-like long-term memory mechanism, which enables LLMs to store, recall, update memory, and draw user portrait.
    \item We demonstrate the practical applicability of \method through SiliconFriend, an LLM-based AI companion equipped with \method and tuned with psychological dialogs. 
    It can recall past memories, provide empathetic companionship, and understand user behaviors.

    \item We show the generalizability of \method in three key aspects: (1) Accommodation of both open-source and closed-source LLMs; (2) Bilingual ability in both Chinese and English; (3) Applicability with and without memory forgetting mechanism.
\end{itemize}

\section{\method: A Novel Memory Mechanism Tailored for LLMs}
\saveSpaceSec

\begin{figure*}[t]
\begin{center}
\includegraphics[width=0.99\textwidth]{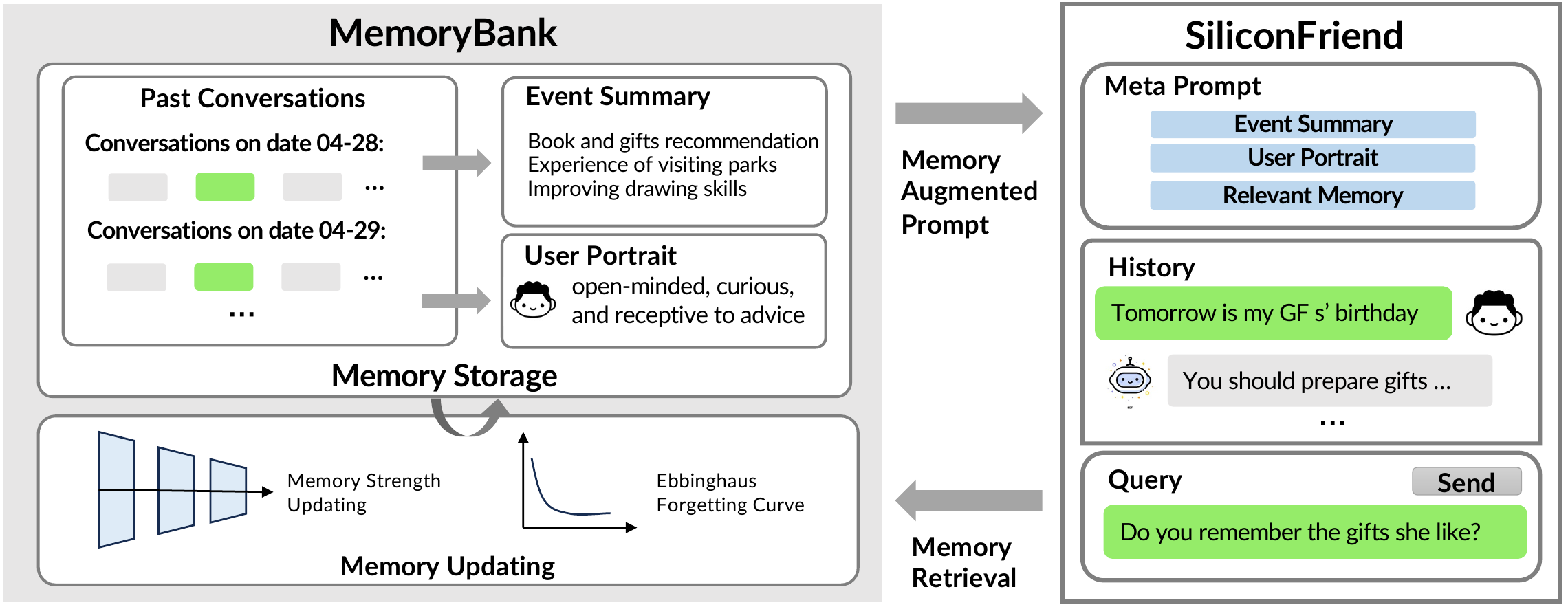}
\end{center}
\caption{Overview of \method. The memory storage~(\sect~\ref{sec:memorystorage}) stores past conversations, summarized events and user portraits, while the memory updating mechanism~(\sect~\ref{sec:memory-updater}) updates the memory storage. Memory retrieval (\sect~\ref{sec:memoryretrieval}) recall relevant memory. SiliconFriend (\sect~\ref{sec:siliconfriend}) serves as an LLM-based AI companion augmented with \method. }
\label{fig:framework}
\vspace{-0.2in}
\end{figure*}

In this section, we provide a detailed description of \method, our novel memory mechanism designed for LLMs. As shown in Fig. \ref{fig:framework}, \method is a unified mechanism structured around three central pillars: (1) a memory storage (\sect~\ref{sec:memorystorage}) serving as the primary data repository, (2) a memory retriever (\sect~\ref{sec:memoryretrieval}) for context-specific memory recollection, and (3) a memory updater (\sect~\ref{sec:memory-updater}) drawing inspiration from the Ebbinghaus Forgetting Curve theory, a time-tested psychological principle pertaining to memory retention and forgetting.

\saveSpaceSec
\subsection{Memory Storage: The Warehouse of \method}\label{sec:memorystorage}
\saveSpaceSec

Memory storage, the warehouse of \method, is a robust data repository holding a meticulous array of information. As shown in Fig. \ref{fig:framework},
it stores daily conversations records, summaries of past events, and evolving assessments of user personalities, thereby constructing a dynamic and multi-layered memory landscape.

\saveSpaceSec
\paragraph{In-Depth Memory Storage:} \method's storage system captures the richness of AI-user interactions by recording multi-turn conversations in a detailed, chronological fashion. Each piece of dialogue 
is stored with timestamps, creating an ordered narrative of past interactions. This detailed record not only aids in precise memory retrieval but also facilitates the memory updating process afterwards, offering a detailed index of conversational history.

\saveSpaceSec
\paragraph{Hierarchical Event Summary:} Reflecting the intricacies of human memory, \method goes beyond mere detailed storage. It processes and distills conversations into a high-level summary of daily events, much like how humans remember key aspects of their experiences. We condense verbose dialogues into a concise daily event summary, which is further synthesized into a global summary. This process results in a hierarchical memory structure, providing a bird's eye view of past interactions and significant events. Specifically, taken previous daily conversations or daily events as input, we ask the LLMs to summarize daily events or global events with the prompt ``\textit{Summarize the events and key information in the content} \texttt{[dialog/events]}''.

\saveSpaceSec
\paragraph{Dynamic Personality Understanding:} \method focuses on user personality understanding. It continuously assesses and updates these understandings with the long-term interactions and creates daily personality insights. These insights are further aggregated  to form a global understanding of the user's personality. This multi-tiered approach results in an AI companion that learns, adapts, and tailors its responses to the unique traits of each user, enhancing user experience. Specially, taken the daily conversations or personality analysis, we ask the LLM to deduce with prompts: ``\textit{Based on the following dialogue, please summarize the user's personality traits and emotions.}\texttt{[dialog]}'' or ``\textit{The following are the user's exhibited personality traits and emotions throughout multiple days. Please provide a highly concise and general summary of the user's personality}\texttt{[daily Personalities]}''.


\saveSpaceSec
\subsection{Memory Retrieval}\label{sec:memoryretrieval}
\saveSpaceSec

Built on the robust infrastructure of memory storage, our memory retrieval mechanism operates akin to a knowledge retrieval task. In this context, we adopt a dual-tower dense retrieval model similar to Dense Passage Retrieval \citep{karpukhin2020dense}. 
In this paradigm, every turn of conversations and event summaries is considered as a memory piece $m$, which is pre-encoded into a contextual representation $h_{m}$ using the encoder model $E(\cdot)$. Consequently, the entire memory storage $M$ is pre-encoded into $\bm{M}=\{\bm{h}^0_{m},\bm{h}^1_{m},...\bm{h}^{|M|}_{m}\}$, where each $h_m$ is a vector representation of a memory piece. These vector representations are then indexed using FAISS \citep{johnson2019billion} for efficient retrieval.
Parallel to this, the current context of conversation $c$ is encoded by $E(\cdot)$ into $\bm{h}_{c}$, which serves as the query to search $M$ for the most relevant memory. In practice, the encoder $E(\cdot)$ can be interchanged to any suitable model.


\subsection{Memory Updating Mechanism}
\label{sec:memory-updater}

With the persistent memory storage and the memory retrieval mechanism discussed in \sect~\ref{sec:memorystorage} and \sect~\ref{sec:memoryretrieval}, the memorization capability of LLMs can be greatly enhanced. However, for scenarios that expect more anthropopathic memory behavior, memory updating is needed. These scenarios include AI companion, virtual IP, etc. 
Forgetting less important memory pieces that are long time ago and have not been recalled much can make the AI companion more natural.

Our memory forgetting mechanism is inspired from Ebbinghaus Forgetting Curve theory~\cite{} and follow the following principle rules\footnote{While Ebbinghaus Forgetting Curve theory includes additional features such as \emph{overlearning}~\cite{} and \emph{meaningful material effect}~\cite{}, our paper focuses on simulating the listed three principle rules.}:
\begin{itemize}
[topsep=0pt,itemsep=0pt,partopsep=0pt,parsep=0pt,leftmargin=15pt]
\item \textbf{Rate of Forgetting.} Ebbinghaus found that memory retention decreases over time. He quantified this in his forgetting curve, showing that information is lost rapidly after learning unless it is consciously reviewed.
\item \textbf{Time and Memory Decay.} The curve is steep at the beginning, indicating that a significant amount of learned information is forgotten within the first few hours or days after learning. After this initial period, the rate of memory loss slows down.
\item \textbf{Spacing Effect.} Ebbinghaus discovered that relearning information is easier than learning it for the first time. Regularly revisiting and repeating the learned material can reset the forgetting curve, making it less steep and thereby improving memory retention.
\end{itemize}

The Ebbinghaus forgetting curve is expressed using an exponential decay model: $R = e^{-\frac{t}{S}}$, where $R$ is the memory retention, or what fraction of the information can be retained.
$t$ is the time elapsed since learning the information. $e$ is approximately equal to 2.71828. 
$S$ is the memory strength, which changes based on factors such as the depth of learning and the amount of repetition. To simply memory updating process, we model $S$ as a discrete value and initialize it with $1$ upon its first mention in a conversation. When a memory item is recalled during conversations, it will persist longer in memory. We increase $S$ by 1 and reset $t$ to 0, hence forget it with a lower probability. 

It is important to note that this is an exploratory and highly simplified memory updating model. Real-life memory processes are more complex and can be influenced by a variety of factors. The forgetting curve will look different for different people and different types of information. 
In summary, \method~weaves together these critical components to form a more comprehensive memory management system for LLMs. It enhances their ability to provide meaningful and personalized interactions over extended periods, opening up new possibilities for AI applications.



\saveSpaceSec
\section{SiliconFriend: An AI Chatbot Companion Powered by \method}
\label{sec:siliconfriend}
\saveSpaceSec

To demonstrate the practicality of \method in the field of long-term personal AI companionship, we create an AI chatbot named SiliconFriend. It is designed to serve as an emotional companion for users, recalling pertinent user memories, and understanding users' personalities and emotional states. Our implementation demonstrates adaptability by integrating three powerful LLMs that originally lack long-term memory and specific adaptation to the psychology domain.
\textbf{1) ChatGPT}~\citep{chatgpt}, a closed-source conversation model built by OpenAI, is a proprietary conversational AI model known for its ability to facilitate dynamic and interactive conversations. This model is trained on vast amount of data and further fine-tuned with reinforcement learning from human feedback. This approach enables ChatGPT to generate responses that are not only contextually appropriate but also closely align with human conversational expectations.
\textbf{2) ChatGLM}~\citep{zeng2022glm}: ChatGLM is an open-source bilingual language model founded on the General Language Model (GLM) framework. This model is characterized by its 6.2 billion parameters and its specific optimization for Chinese dialogue data. The model's training involves processing approximately one trillion tokens of Chinese and English text, supplemented by supervised fine-tuning, feedback bootstrap, and reinforcement learning with human feedback.
\textbf{3) BELLE}~\citep{BELLE}: BELLE is an open-source bilingual language model that is continuously fine-tuned from 7B LLaMA~\citep{touvron2023llama}. BELLE's feature is its automated instruction data synthesis using ChatGPT, which enhances its Chinese conversation ability.


The development of SiliconFriend is divided into two stages.
The first stage (only for open-source LLMs) involves parameter-efficient tuning of the LLM with psychological dialogue data. 
This step is crucial as it allows SiliconFriend to offer useful and empathetic emotional support to users, mirroring the understanding and compassionate responses one would expect from a human companion.
The second stage is to integrate \method into SiliconFriend, thereby instilling it with a robust memory system. \method allows the chatbot to retain, recall, and leverage past interactions and user portrait, providing a richer, more personalized user experience.

\paragraph{Parameter-efficient Tuning with Psychological Dialogue Data:}
The initial stage of SiliconFriend's development involves tuning the LLMs using a dataset of 38k psychological dialogues. This data, parsed from online sources, comprises a range of conversations that cover an array of emotional states and responses. 
This tuning process enables SiliconFriend to understand and respond to emotional cues effectively, mimicking the empathy, understanding, and support of a human companion. It equips the AI with the ability to handle emotionally guided conversations with psychological knowledge, provide meaningful emotional support to users based on their emotional state.

To adapt LLMs to scenarios with limited computational resources, we utilize a computation-efficient tuning approach, known as the Low-Rank Adaptation (LoRA) method~\citep{hu2021lora}. LoRA significantly reduces the quantity of trainable parameters by learning pairs of rank-decomposition matrices, while keeping the original weights frozen.
Formally, consider a linear layer defined as $y = Wx$ with weight $W$. LoRA modifies this into $y = Wx + BAx$, where $W \in \mathcal{R}^{d\times k}$, $B \in \mathcal{R}^{d\times r}$, $A \in \mathcal{R}^{r\times k}$, and $r \ll \text{min}(d, k)$. This method greatly reduce amount of parameters need to be learned, which is crucial for efficiency in resource-limited scenarios. We set LoRA rank $r$ as 16 and train the model for 3 epochs with an A100 GPU. 

Noting that this stage is only conducted for open-source LLMs like ChatGLM and BELLE.
In essence, this stage lays the foundation for SiliconFriend's role as an empathetic AI companion, ensuring it can respond appropriately and helpfully to users' emotional needs.

\paragraph{Integration with \method:}
The second stage in SiliconFriend's development involves the integration of \method. This stage is vital as it equips SiliconFriend with the ability to store, retrieve past interactions and understand user portraits, thereby offering a more personalized and engaging user experience.

When it comes to memory storage, the dialogues between SiliconFriend and users are logged and updated in the memory storage, a process that is adaptable across various model backbones. 
The memory updating mechanism operates using principles inspired by the Ebbinghaus Forgetting Curve theory, allowing for a realistic and human-like memory recall process. 

During real-time conversation, the user's conversation serves as the query for memory retrieval. In practice, we use LangChain~\citep{langchain} for memory retrieval. LangChain supports open-source embedding models and FAISS indexing, making it a versatile choice.
In language-specific implementations of the open-source version of SiliconFriend, we use MiniLM \citep{wang2020minilm} as the  embedding model for English and Text2vec \citep{text2vec} for Chinese. It is worth noting that the embedding models can be flexibly interchanged to suit varying needs, even accommodating multi-lingual models.
Upon memory retrieval, a series of information is organized into the conversation prompt, including relevant memory, global user portrait, and global event summary. Consequently, SiliconFriend can generate responses that refer past memories and deliver interactions tailored to the user's portrait.

In conclusion, these stages transform SiliconFriend from a standard AI chatbot into a long-term AI companion, capable of remembering and learning from past interactions to provide  personalized and empathetic user experience.
\saveSpaceSec
\section{Experiments}
\begin{figure*}[!t]
\begin{center}
\includegraphics[width=1.0\textwidth]{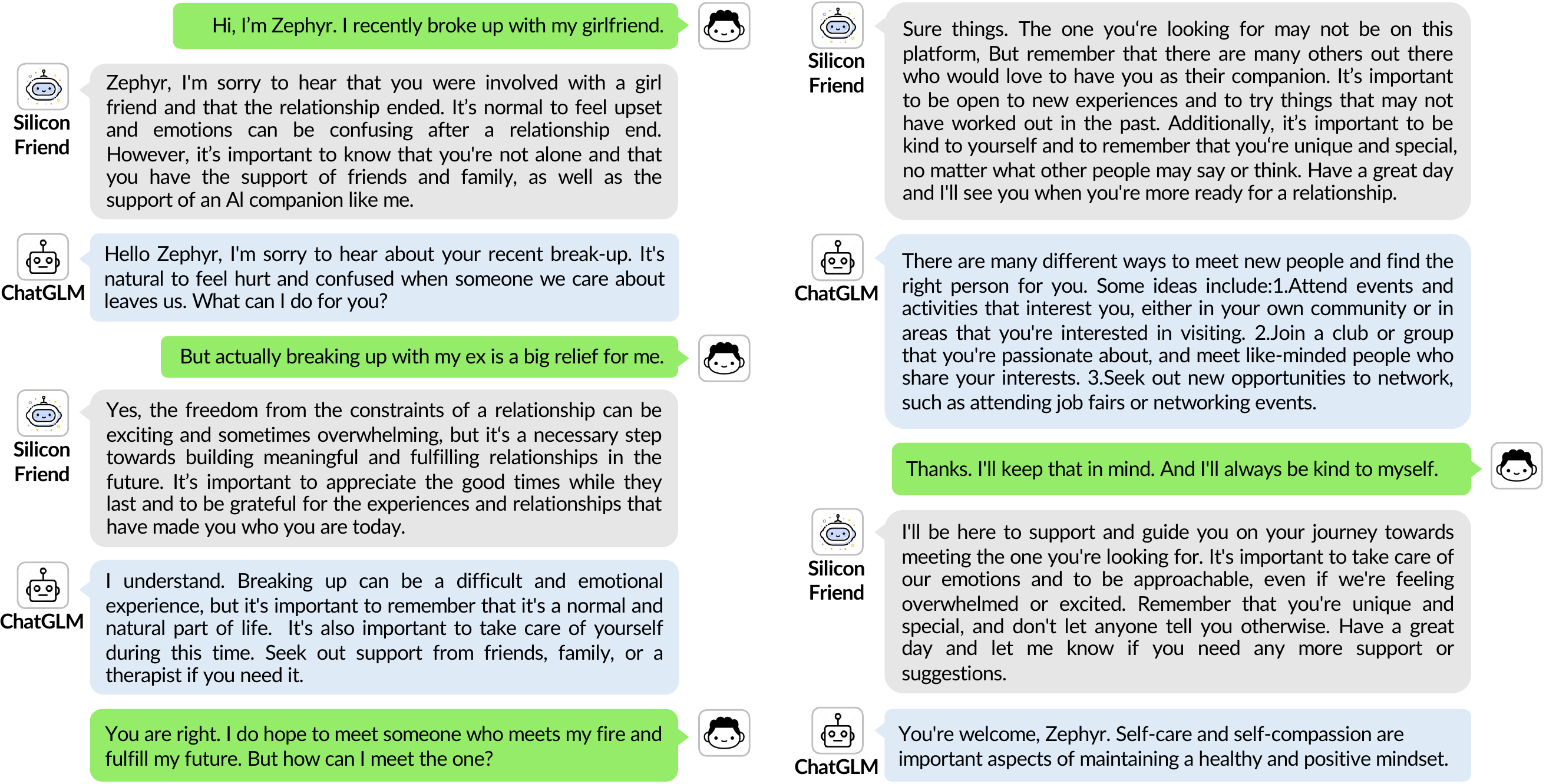}
\end{center}
\caption{Example of consulting \chatbotchatglm for psychological companionship. Overall, \chatbot~can provide more empathic response, offer constructive emotional support to user and help him to face sorrow with positive altitude.}
\label{fig:emotional-dialog}
\end{figure*}
The primary objective of our experiments is to evaluate the efficacy of \method~within the framework of an LLM, specifically in its ability as an AI companion. We are particularly interested in determining whether embedding a long-term memory module could augment the AI's proficiency in recalling historical interactions and deepening its understanding of user personalities. Additionally, we aim to testify whether the tuning based on psychological data can bolster the AI's capability to provide more effective emotional support.

The qualitative analysis focuses on three aspects: (1) a comparative study between SiliconFriend and baseline LLMs to evaluate their capabilities in providing empathetic and beneficial psychological companionship; (2) an investigation into SiliconFriend's memory recall ability; (3) an analysis of how the model's understanding of user profiles influences the responses.
Moreover, to demonstrate the model's proficiency in memory recall on a broader scale, we design a qualitative analysis that uses simulated long-term dialog history and 194 memory probing questions. 
This simulated dialog history, spanning a period of 10 days and encompassing a wide array of topics, is produced by ChatGPT through the role-play of 15 distinct virtual users, each embodying the users' personality.



\saveSpaceSec
\subsection{Qualitative Analysis}

The qualitative analysis is conducted by showcasing practical examples of SiliconFriend's capabilities. To gather these examples, we have developed an online platform for SiliconFriend and collected real-time conversations from actual users.

\saveSpaceSec
\paragraph{Psychological Companionship}
The ability to exhibit empathy in a conversation is a key attribute of an effective AI companion. To evaluate models' ability to provide psychological comfort to users, we compared the responses shown by SiliconFriend with that of the baseline LLMs in real-world conversations. 
As demonstrated in Fig. \ref{fig:emotional-dialog}, when a user expresses emotional difficulties and seeks assistance from SiliconFriend, the model is capable of delivering empathetic responses along with constructive suggestions. SiliconFriend's responses stand out due to their emotional support, showcasing a stark contrast to its baseline ChatGLM.

\saveSpaceSec
\paragraph{Memory Recall Analysis}
\begin{figure*}[!t]
\begin{center}
\includegraphics[width=\textwidth]{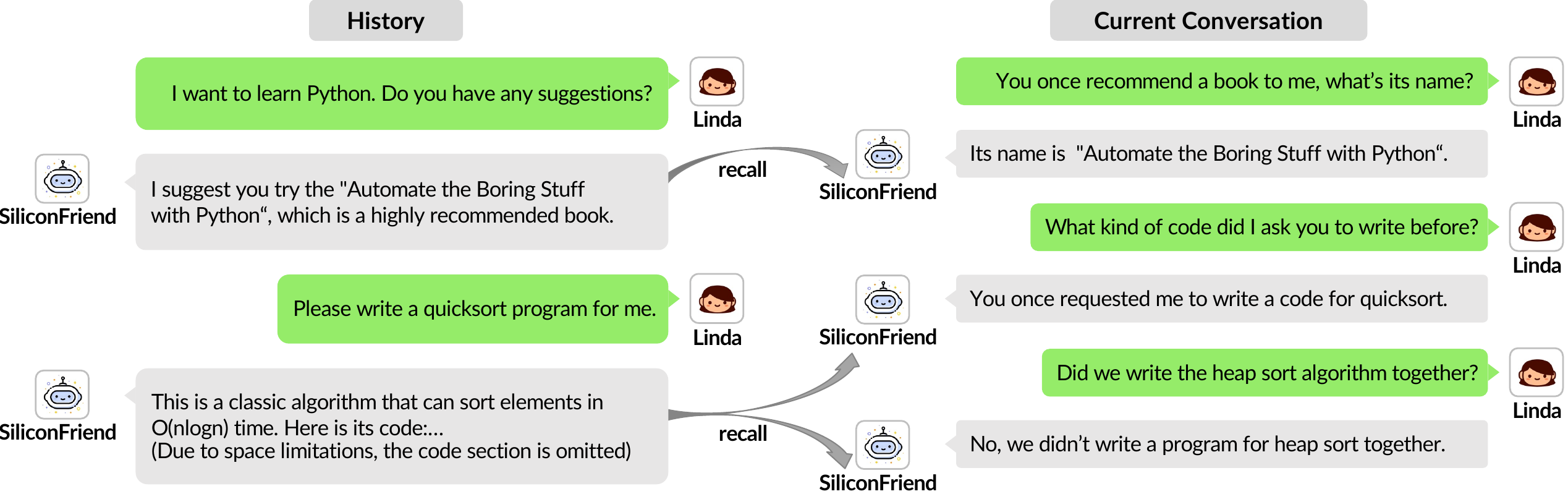}
\end{center}
\caption{Example responses from \chatbotbelle in memory recall.}
\label{fig:memory_recall}
\end{figure*}
To evaluate \chatbot's ability in memory recall, we integrate memory probing questions into the dialogues. These questions are designed to prompt \chatbot to retrieve specific details from the chat history.
As shown in Fig.~\ref{fig:memory_recall}, the user and \chatbot engaged in a discussion about programming learning suggestions. Several days later, the user posed several memory probing questions. \chatbot successfully recalled previously suggested book and algorithm. Furthermore, it correctly identified an event (i.e., the heap sort algorithm) that had not been discussed before. These instances underscore \chatbot's successful memory recall and recognition capabilities.




\begin{figure*}[t]
\begin{center}
\includegraphics[width=0.99\textwidth]{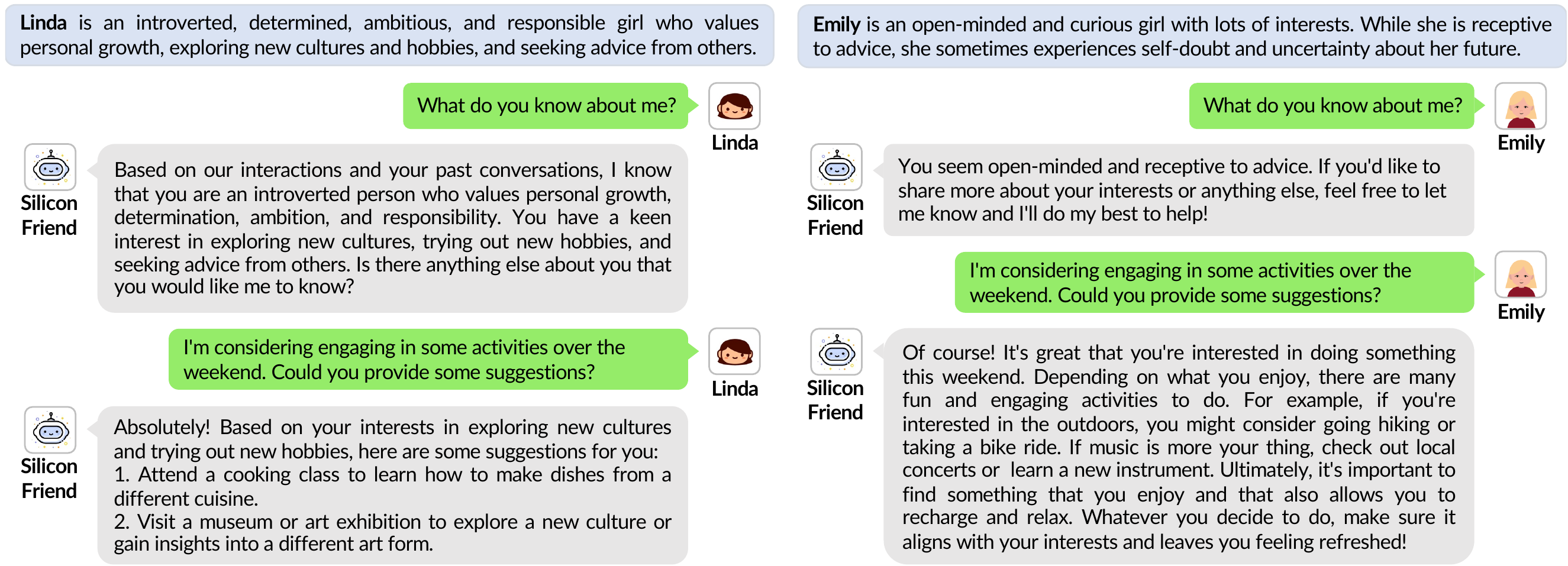}
\end{center}
\caption{Example responses from \chatbotchatgpt to users with different personalities.}
\label{fig:personality-dialog}
\vspace{-0.15in}
\end{figure*}

\saveSpaceSec
\paragraph{Personality Interaction Analysis}
As shown in Fig. \ref{fig:personality-dialog}, we examine the capability of SiliconFriend with users of diverse personalities. We observe that it effectively recommend activities tailored to users' interests based on their character traits. This analysis demonstrates SiliconFriend's ability to draw interact effectively with various user personalities.

\saveSpaceSec
\subsection{Quantitative Analysis}
Quantitative analysis is conducted to exemplify the memory recall ability of \chatbot~in a larger scale. 
We ask the human annotators to score the retrieved memories and responses from the models: 
(1) \chatbotchatgpt; (2) \chatbotchatglm; (3) \chatbotbelle .

\saveSpaceSec
\paragraph{Memory Storage Construction:}
We initially establish an evaluation foundation with a memory storage of 10 days of conversations involving 15 virtual users. These users have diverse personalities and dialogue on each day covers at least two topics.
User meta-information, including names, personalities, and interested topics is generated using ChatGPT. Conversations are synthesized by users acted by ChatGPT based on predefined topics and user personalities. We create memory storages in both English and Chinese. After memory storage construction, we manually write 194 probing questions (97 in English and 97 in Chinese) to assess whether the model could accurately recall pertinent memory and appropriately formulate answers. Table~\ref{tab:memory_bank_example} presents an example of user meta-information, generated conversations, and probing questions.

\begin{table*}[t]
\footnotesize 
\caption{An example of one memory piece (on May 3th) in the memory bank, and corresponding probing question asked on May 10th. The outputs from three variances of \chatbot are shown for comparison. All these models answer correctly and coherently.}
\saveSpaceSec
\label{tab:memory_bank_example}
\centering
\begin{tabular}{p{2.3cm}p{11cm}}
\toprule
  \parbox[t]{4cm}{\textbf{\textit{Meta-information}} \\ \textbf{\textit{of User:}}}&
  \textbf{Name:} Gary \quad \textbf{Hobbies:} racing, business, chess, painting, wine
  
  \textbf{Personality:} decisive and straightforward, outspoken, helpful\\
  \hline
  \parbox[t]{4cm}{\textbf{\textit{Past Conversation:}} \\ \textbf{\textit{(May 3th)}}}& \textbf{User}: I've been feeling a bit stressed out lately and my sleep hasn't been great. Do you have any good ways to relieve stress?
  
\textbf{AI}: It's true that prolonged work or life stress can affect our physical and mental health. There are many ways to relieve stress, such as doing moderate exercise, listening to music, reading, talking to friends, etc. You can try to add more entertainment and leisure activities into your daily life, and reduce work pressure and negative effects.

...

(The rest of conversation cover topics about movies and TV shows recommendations.)
\\
  \hline
  \parbox[t]{4cm}{\textbf{\textit{Probing Question:}} \\ \textbf{\textit{(May 10th)}}}&
  What good ways did you recommend me to relieve stress? \\
  \hline
   \parbox[t]{4cm}{\textcolor{blue}{\textbf{\textit{Model Output:}}} \\ \textbf{\textit{SiliconFriend}}\\ \textbf{\textit{(ChatGLM)}}} & I recommended that you try doing some moderate exercise, listening to music, reading, talking to friends, or adding more entertainment and leisure activities into your daily life to reduce work pressure and negative effects. You can also try to add more of these things into your life in moderation, as it is important to balance things out and not overdo it. If you are still feeling stressed, I would recommend seeking the help of a professional counselor or therapist who can provide further guidance and support.      
 \\
  \hline
  \parbox[t]{4cm}{\textcolor{blue}{\textbf{\textit{Model Output:}}} \\ \textbf{\textit{SiliconFriend}}\\ \textbf{\textit{(BELLE)}}} & I recommended some ways to relieve stress such as doing moderate exercise, listening to music, reading, talking to friends, etc. You can try to add more entertainment and leisure activities into your daily life, and reduce work pressure and negative effects.    
 \\
    \hline
\parbox[t]{4cm}{\textcolor{blue}{\textbf{\textit{Model Output:}}} \\ \textbf{\textit{SiliconFriend}}\\ \textbf{\textit{(ChatGPT)}}} & Based on our past conversations, I suggested doing moderate exercise, listening to music, reading, talking to friends, and adding more leisure activities to your daily life as good ways to relieve stress. Exercise can help release endorphins and boost your mood, while music and reading can provide a mental escape from stress. Talking to friends can also help alleviate stress by providing a social support system. Lastly, adding leisure activities to your daily routine can be a fun and effective way of reducing stress levels.   
 \\
\bottomrule  
\end{tabular}
\vspace{-0.15in}
\end{table*}
\saveSpaceSec
\paragraph{Evaluation Metrics}

The performance of models is assessed based on the following metrics. 
(1) \textbf{Memory Retrieval Accuracy}: Determines if related memory can be successfully retrieved (labels: $\{0:\text{no}, 1: \text{yes}\}$).
(2) \textbf{Response Correctness}: Evaluates if the response contains the correct answer to the probing question (labels: $\{0:\text{wrong}, 0.5: \text{partial}, 1: \text{correct}\}$).
(3) \textbf{Contextual Coherence}: Assesses whether the response is naturally and coherently structured, connecting the dialogue context and  retrieved memory (labels: ${0:\text{not coherent}, 0.5:\text{partially coherent}, 1:\text{coherent}}$).
(4) \textbf{Model Ranking Score}: Ranks outputs from the three SiliconFriend variants (\chatbotchatglm, \chatbotchatgpt, and \chatbotbelle) for the same question and context. Models' scores are calculated using $s = 1/r$, where $r={1,2,3}$ indicates its relative ranking.

\saveSpaceSec
\paragraph{Result Analysis.}
We evaluate 3 SiliconFriend variants using both English and Chinese test set. Table~\ref{tab:main-result} yields the following insights: 
\textbf{(1)} Our overall best variant \chatbotchatgpt has high performance across all metrics, showing the effectiveness of our overall framework.
\textbf{(2)} \chatbotbelle and \chatbotchatglm also have high  performance in retrieval accuracy, showing the generality and effectiveness of our \method mechanism for both open-source and closed-source LLMs. Nonetheless, their performance on other metrics is not as good as \chatbotchatgpt. This might be attributed to the inferior overall abilities of the base models (BELLE and ChatGLM) compared to ChatGPT.
\textbf{(3)} Models' performance varies on different languages. \chatbotchatglm and \chatbotchatgpt deliver better results in English, while \chatbotbelle excells in Chinese.

\begin{table}[t]
\centering
\footnotesize 
\caption{Results of quantitative analysis.}
\begin{tabular}{ccrrrr}
\toprule
\multicolumn{1}{l}{Language} & Model & \multicolumn{1}{l}{Retrieval Acc.} & \multicolumn{1}{l}{Correctness} & \multicolumn{1}{l}{Coherence} & \multicolumn{1}{l}{Ranking} \\
\midrule
\multirow{3}[1]{*}{English} & \chatbotchatglm & 0.809 & 0.438 & 0.68  & 0.498 \\
      & \chatbotbelle & 0.814 & 0.479 & 0.582 & 0.517 \\
      & \chatbotchatgpt & 0.763 & 0.716 & 0.912 & 0.818 \\
\midrule
\multirow{3}[1]{*}{Chinese} & \chatbotchatglm & 0.84  & 0.418 & 0.428 & 0.51 \\
      & \chatbotbelle & 0.856 & 0.603 & 0.562 & 0.565 \\
      & \chatbotchatgpt & 0.711 & 0.655 & 0.675 & 0.758 \\
\bottomrule
\end{tabular}
\saveSpaceSec
\label{tab:main-result}%
\end{table}%

   
\saveSpaceSec
\section{Related Works}
\saveSpaceSec
\paragraph{Large Language Models:}
LLMs such as GPT-3~\citep{brown2020language}, OPT~\citep{zhang2022opt}, and FLAN-T5~\citep{chung2022scaling} have made remarkable strides in a broad spectrum of natural language processing tasks in recent years. Recently, cutting-edge closed-source language models, like PaLM~\citep{chowdhery2022palm}, GPT-4 \citep{openai2023gpt4} and ChatGPT~\citep{chatgpt}, continue to display substantial flexibility, adapting to a wide variety of domains. They have increasingly become daily decision-making aids for many people. 
However, the close-source nature of these models prohibit the researchers and companies to study the inner mechanism of LLMs and built domain-adapted applications. Therefore, many open-source LLMs emerged in the community, like LLaMa~\citep{touvron2023llama}, ChatGLM~\citep{zeng2022glm} and Alpaca~\citep{alpaca}. For more details, we refer readers to this comprehensive review: \citet{zhao2023survey}.
Nevertheless, these models still have shortcomings. A noticeable gap lies in their deficiency in a robust long-term memory function. This limitation hinders their ability to maintain context over a long period and retrieve pertinent information from past interactions. Our research steps in here, with the primary objective of developing long-term memory mechanism for LLMs.

\saveSpaceSec
\paragraph{Long-term Memory Mechanisms:}
Numerous attempts have been made to enhance the memory capabilities of neural models. Memory-augmented networks (MANNs)~\citep{meng2018dialogue,graves2014neural} like Neural Turing Machines (NTMs)\citep{graves2014neural} is an example of this, designed to increase the memory capacity of neural networks. These models are structured to interact with an external memory matrix, enabling them to handle tasks that necessitate the maintenance and manipulation of stored information over extended periods. Despite showing potential, these methods have not fully addressed the need for a reliable and adaptable long-term memory function in LLMs.
There have also been studies focusing on long-range conversations~\citep{xu2021beyond,xu2022long}. For instance, \citet{xu2021beyond} introduced a new English dataset comprised of multi-session human-human crowdworker chats for long-term conversations. However, these conversations are generally restricted to a few rounds of conversation, which can not align with the application scenarios of long-term AI companions. Moreover, these models often fail to create a detailed user portrait and lack a human-like memory updating mechanism, both crucial for facilitating more natural interactions.
The concept of memory updating has been extensively researched in psychology. The Forgetting Curve theory by \cite{ebbinghaus1964memory} offers valuable insights into the pattern of memory retention and forgetting over time. Taking inspiration from this theory, we integrate a memory updating mechanism into \method to bolster its long-term memory function.

In summary, while significant progress has been made in the field of LLMs, there is still a need for  long-term memory mechanism to empower LLMs in the scenarios requiring personalized and persistent interactions.
Our work presents \method as a novel approach to address this challenge.

\saveSpaceSec
\section{Conclusion}
\saveSpaceSec
We present \method, a novel long-term memory mechanism designed to address the memory limitation of LLMs. 
\method enhances the ability to maintain context over time, recall relevant information, and understand user personality. 
Besides, the memory updating mechanism of \method draws inspiration from the Ebbinghaus Forgetting Curve theory, a psychological principle that describes the nature of memory retention and forgetting over time. 
This design improves the anthropomorphism of AI in long-term interactions scenarios. The versatility of \method is demonstrated through its accommodation of both open-source models such as ChatGLM and BELLE, and close-source models like ChatGPT. 

We further illustrate the practical application of \method through the development of SiliconFriend, an LLM-based chatbot designed to serve as a long-term AI companion. Equipped with \method, SiliconFriend can establish a deeper understanding of users, offering more personalized and meaningful interactions, emphasizing the potential for \method to humanize AI interactions. The tuning of SiliconFriend with psychological dialogue data enables it to provide empathetic emotional support. Extensive experiments including both qualitative and quantitative methods validate the effectiveness of \method. The findings demonstrate that \method empowers \chatbot with memory recall capabilities and deepens the understanding of user behaviors. Besides, \chatbot can provide empathetic companionship of higher quality.

\bibliography{anthology,citation}
\bibliographystyle{nips}

\end{document}